\documentclass[lettersize,journal]{IEEEtran}
\usepackage{amsmath,amsfonts}
\usepackage{algorithmic}
\usepackage{algorithm}
\usepackage{array}
\usepackage[caption=false,font=normalsize,labelfont=sf,textfont=sf]{subfig}
\usepackage{textcomp}
\usepackage{stfloats}
\usepackage{url}
\usepackage{verbatim}
\usepackage{graphicx}
\usepackage{cite}
\usepackage{longtable}
\usepackage{multirow}
\usepackage{booktabs}

\usepackage{pifont}
\usepackage{tikz}
\usetikzlibrary{shadows.blur}  
\usepackage[most]{tcolorbox}
\tcbuselibrary{skins}         
\usepackage{graphicx}         
\tcbuselibrary{skins,breakable} 
\usepackage{hyperref}

\hypersetup{
    colorlinks=true,
    citecolor=cyan!85!teal,    
    linkcolor=cyan!85!teal,    
    urlcolor=cyan!85!teal     
}

\tcbset{
  enhanced,                       
  colframe=teal!70!black,        
  colback=white!95!teal,          
  colbacktitle=teal!60!black,    
  coltitle=white,                 
  boxrule=0.8mm,                  
  arc=2mm,                       
  drop shadow=teal!50!black,      
  left=7mm, right=7mm, top=6mm, bottom=6mm, 
  fonttitle=\bfseries\normalsize\sffamily, 
  before skip=12pt,              
  after skip=12pt,               
  attach boxed title to top center={yshift=-2mm},
  boxed title style={
    colframe=teal!70!black,       
    colback=teal!60!black,        
    boxrule=0.8mm,                
    rounded corners=all,         
    arc=2mm                       
  }
}

\newcommand{\pnum}[1]{%
    \tikz[baseline=(char.base)]{
        \node[
            rounded corners=3pt,   
            fill=teal,       
            text=white,              
            inner sep=2pt,           
            minimum size=1em,      
            font=\footnotesize\bfseries    
        ] (char) {#1};
    }
}

\begin{document}

\title{Emoti-Attack: Zero-Perturbation Adversarial Attacks on NLP Systems via Emoji Sequences}

\author{Yangshijie Zhang\\
        Lanzhou University, Lanzhou, China\\
        Email: zhangyshj2023@lzu.edu.cn
        
}

\IEEEpubid{0000--0000/00\$00.00~\copyright~2025 IEEE}

\maketitle

\begin{abstract}
Deep neural networks (DNNs) have achieved remarkable success in the field of natural language processing (NLP), leading to widely recognized applications such as ChatGPT. However, the vulnerability of these models to adversarial attacks remains a significant concern. Unlike continuous domains like images, text exists in a discrete space, making even minor alterations at the sentence, word, or character level easily perceptible to humans. This inherent discreteness also complicates the use of conventional optimization techniques, as text is non-differentiable. Previous research on adversarial attacks in text has focused on character-level, word-level, sentence-level, and multi-level approaches, all of which suffer from inefficiency or perceptibility issues due to the need for multiple queries or significant semantic shifts.

In this work, we introduce a novel adversarial attack method, Emoji-Attack, which leverages the manipulation of emojis to create subtle, yet effective, perturbations. Unlike character- and word-level strategies, Emoji-Attack targets emojis as a distinct layer of attack, resulting in less noticeable changes with minimal disruption to the text. This approach has been largely unexplored in previous research, which typically focuses on emoji insertion as an extension of character-level attacks. Our experiments demonstrate that Emoji-Attack achieves strong attack performance on both large and small models, making it a promising technique for enhancing adversarial robustness in NLP systems.
\end{abstract}


\section{Introduction}
\IEEEPARstart{D}{eep} 
 neural networks (DNNs) have garnered significant achievements within the NLP field, giving rise to renowned applications like ChatGPT . Nevertheless, it is crucial to focus on the vulnerability of these NLP models to adversarial attacks for their protection.  Contrary to the continuous nature of the image domain, text exists in a discrete space. This implies that any alterations, whether at the sentence, word, or even character level, are readily noticeable to humans, making it challenging to create unnoticeable perturbations.

Furthermore, the discrete nature of text data renders it non-differentiable, leading to the ineffectiveness of conventional optimization techniques. Studies on adversarial textual attacks have been broadly categorized into character-level, word-level, sentence-level, and multi-level approaches . At the character level, attacks involve altering words by means of insertion, omission, misspelling, substitution, or transposition of characters. Such modifications are readily identifiable by people. Yet, character-level attacks necessitate numerous queries to ascertain the specific characters to target. Word-level tactics involve modifying text by inserting, removing, or substituting keywords. Like their character-level counterparts, these attacks also demand repeated queries to identify the target words. Both attacks are iterative, potentially leading to inefficiency in practical NLP use cases due to the need for multiple queries per iteration. Sentence-level attacks entail altering the entire text by rephrasing or appending nonsensical sentences to the original text. The substantial alterations caused by sentence-level attacks might occasionally alter the fundamental meaning of the text. Perturbations at the character, word, and sentence levels can result in significant semantic shifts from the original content and are readily discernible by human observers.

We propose a groundbreaking technique known as emoji-level attacks within textual adversarial methods. This approach, which diverges from character- and word-level attacks, targets the manipulation of emoji, offering a less noticeable, and less-word-perturbation. Studies to date have not adequately explored the potential of emoji in adversarial contexts, with some research merely toying with emoji insertion as an extension of character-level strategies.

Our innovation lies in treating emoji modifications as a distinct layer of attack on par with character, word, and sentence-level approaches.

Our contributions are as following:

(1) We propose Emoji-Attack , a novel type of adversarial attack that utilizes seemingly harmless or even playful emoticons to manipulate NLP systems.

(2) Emoji-Attack demonstrates strong attack performance on both large and small models.

\section{Problem Formulation}
\enlargethispage{-1\baselineskip}

\subsection{Zero-Word-Perturbation Attack Framework}

Contemporary adversarial attack methods predominantly rely on direct text modifications, inevitably compromising semantic integrity and triggering detection mechanisms~\cite{yang2023semantic}. Building upon insights from~\cite{ma2024security} demonstrating the significant influence of emojis on NLP model behavior, we propose a novel zero-word-perturbation adversarial attack framework. Our approach achieves adversarial effects through strategic emoji sequence placement while preserving complete textual integrity.

For a given text $x$ and emoji sequences $s, s' \in \mathcal{S}(\mathcal{E})$, we formalize the sequence concatenation operation:

\begin{equation}
\label{eq:concat_operation}
s \oplus x \oplus s' = \text{cat}(s) \cdot x \cdot \text{cat}(s'),
\end{equation}
where $\text{cat}$ denotes the sequence concatenation operation, with $s$ and $s'$ representing the prefix and suffix emoji sequences, respectively.

Given a target classification model $f_{\text{tgt}}: \mathcal{X} \to \mathcal{Y}$ and an input text $x \in \mathcal{X}$, our framework identifies emoji sequences $s, s' \in \mathcal{S}(\mathcal{E})$ satisfying:

\begin{equation}
\label{eq:zero_perturb}
f_{\text{tgt}}(s \oplus x \oplus s') \neq f_{\text{tgt}}(x).
\end{equation}

The framework incorporates two fundamental constraints. First, ~\cite{gupta2023emoji} establishes the importance of emotional consistency between emoji sequences and the original text:

\begin{equation}
\label{eq:emotion_constraint}
f_{\text{sen}}(s) = f_{\text{sen}}(s') = f_{\text{sen}}(x).
\end{equation}

Additionally, we impose length constraints: $l_{\text{min}} \leq |s|, |s'| \leq l_{\text{max}}$.

The optimization objective maximizes prediction divergence while maintaining these constraints:

\begin{equation}
\label{eq:opt_objective}
\max_{s,s' \in \mathcal{S}(\mathcal{E})} \mathcal{L}(s \oplus x \oplus s', y),
\end{equation}
where the loss function quantifies the prediction divergence:

\begin{equation}
\small
\label{eq:loss_function}
\mathcal{L}(s \oplus x \oplus s', y) = \log p_{\text{tgt}}(\hat{y}|s \oplus x \oplus s') - \log p_{\text{tgt}}(y|s \oplus x \oplus s').
\end{equation}

Here, $y$ represents the original label, $\hat{y}$ denotes the highest-confidence incorrect label, and $p_{\text{tgt}}(\cdot|\cdot)$ indicates the model's prediction probability.

To evaluate attack stealthiness comprehensively, we introduce a metric function $\eta: \mathcal{X} \times \mathcal{S}(\mathcal{E}) \times \mathcal{S}(\mathcal{E}) \to [0,1]$~\cite{cheng2025strongly}:

\begin{equation}
\label{eq:stealthiness}
\eta(x,s,s') = \alpha \cdot \delta(x,s,s') + (1-\alpha) \cdot \gamma(|s|+|s'|),
\end{equation}
where $\delta(x,s,s')$ assesses sentiment consistency across components, and $\gamma(l)$ implements a length-based penalty:

\begin{equation}
\label{eq:length_penalty}
\gamma(l) = \max(0, 1-\frac{l-l_{\text{min}}}{l_{\text{max}}-l_{\text{min}}}).
\end{equation}

This formalization establishes a novel paradigm for adversarial attacks, achieving effective model manipulation while maintaining perfect textual integrity through strategically positioned emoji sequences.

\subsection{Emoji Sequence Space}

To formalize our zero-word-perturbation attack methodology within a rigorous mathematical framework, we construct a comprehensive emoji sequence space encompassing both standard Unicode emoji characters (e.g., grinning-face, fire) and ASCII-based emoticons (e.g., ":)", "QaQ", ";-P"). This dual-modality representation space enables more expressive adversarial sequence generation while maintaining naturalness in the resultant perturbations. ~\cite{weissman2024compositional} establishes fundamental properties of emoji sequence composition, providing theoretical foundations for our construction.

Let $\mathcal{E}$ denote the finite set of all available emojis. We formally define the emoji sequence space $\mathcal{S}(\mathcal{E})$ as:

\begin{equation}
\label{eq:sequence_space}
\mathcal{S}(\mathcal{E}) = \{s = (e_1,\ldots,e_l) | e_i \in \mathcal{E}, l_{\text{min}} \leq l \leq l_{\text{max}}\}.
\end{equation}

To ensure emotional consistency in sequence construction, we introduce sentiment-specific subspaces. For each sentiment label $y \in \mathcal{Y}$, we define its corresponding sequence subspace $\mathcal{S}_y$ as:

\begin{equation}
\label{eq:sentiment_subspace}
\mathcal{S}_y = \{s \in \mathcal{S}(\mathcal{E}) | f_{\text{sen}}(s) = y\}.
\end{equation}

A fundamental property of our constructed sequence space lies in its completeness. Specifically, for any input text $x \in \mathcal{X}$ and target label $y \in \mathcal{Y}$, there exists at least one emoji sequence $s \in \mathcal{S}(\mathcal{E})$ that simultaneously satisfies both emotional consistency and adversarial effectiveness:

\begin{equation}
\label{eq:completeness}
(f_{\text{sen}}(s) = f_{\text{sen}}(x)) \land (f_{\text{tgt}}(s \oplus x \oplus s') \neq f_{\text{tgt}}(x)).
\end{equation}

The existence of such sequences emerges from the rich combinatorial structure of our emoji sequence space. For sequences of length $l$, the space admits $|\mathcal{E}|^l$ possible combinations, providing substantial flexibility in sequence construction while maintaining semantic naturalness. This theoretical foundation, supported by the compositional properties established in ~\cite{weissman2024compositional}, ensures both the expressiveness and practical applicability of our framework.

\subsection{Emotional Consistency Framework}

A critical component of our zero-word-perturbation attack framework lies in maintaining emotional consistency between the injected emoji sequences and the original text. Recent advances in multimodal emotion analysis~\cite{bao2024research} demonstrate the significance of such alignment in ensuring attack imperceptibility. We present a rigorous mathematical framework for quantifying and ensuring this crucial consistency property.

Let $f_{\text{sen}}: \mathcal{S}(\mathcal{E}) \to \mathcal{Y}$ denote a mapping function that associates emoji sequences with sentiment labels. We formalize the sentiment consistency evaluation through a binary function $\delta: \mathcal{X} \times \mathcal{S}(\mathcal{E}) \times \mathcal{S}(\mathcal{E}) \to \{0,1\}$:

\begin{equation}
\label{eq:sentiment_consistency}
\delta(x,s,s') = \mathbf{1}(f_{\text{sen}}(s) = f_{\text{sen}}(s') = f_{\text{sen}}(x)).
\end{equation}

To establish a comprehensive evaluation framework for generated sequences, we integrate this consistency measure into a broader stealthiness metric. The resultant function $\eta: \mathcal{X} \times \mathcal{S}(\mathcal{E}) \times \mathcal{S}(\mathcal{E}) \to [0,1]$ synthesizes emotional alignment with sequence efficiency:

\begin{equation}
\label{eq:comprehensive_metric}
\eta(x,s,s') = \alpha \cdot \delta(x,s,s') + (1-\alpha) \cdot \gamma(|s|+|s'|),
\end{equation}
where the parameter $\alpha \in [0,1]$ mediates the trade-off between sentiment consistency preservation and sequence length optimization.

Our framework provides rigorous theoretical guarantees regarding the existence of adversarial sequences that simultaneously achieve attack effectiveness and maintain emotional consistency. For any input text $x$ and an arbitrarily small positive constant $\epsilon > 0$:

\begin{equation}
\small
\label{eq:guarantee}
\exists s,s' \in \mathcal{S}(\mathcal{E}): f_{\text{tgt}}(s \oplus x \oplus s') \neq f_{\text{tgt}}(x) \land \eta(x,s,s') \geq 1-\epsilon.
\end{equation}

This fundamental guarantee emerges from two key properties: \pnum{1} the rich expressiveness of the constructed emoji sequence space, and \pnum{2} the careful design of our consistency evaluation mechanism. The guarantee ensures the theoretical feasibility of generating adversarial examples that achieve both attack success and high stealthiness under our comprehensive metric.

\section{Methodology}

\subsection{Two-Phase Learning Framework}

The efficient identification of optimal emoji sequences within the vast search space presents significant computational challenges. ~\cite{tirumala2022behavior} demonstrates that direct optimization through reinforcement learning often converges to suboptimal solutions due to insufficient semantic prior knowledge. Drawing from advances in hybrid learning approaches~\cite{wang2019improved}, we develop a two-phase learning framework that systematically combines supervised pretraining with reinforcement learning optimization.

The first phase employs supervised learning, leveraging an auxiliary sentiment analysis model $f_{\text{sen}}$ to establish fundamental semantic mappings. The optimization objective is formalized as:

\begin{equation}
\label{eq:supervised_loss}
\mathcal{L}_{\text{sup}}(\theta) = -\frac{1}{|\mathcal{D}|} \sum_{(x,s) \in \mathcal{D}} \sum_{t=1}^{l} \log \pi_{\theta}(s_t|s_{1:t-1},x,l),
\end{equation}
where $\pi_{\theta}$ denotes the parameterized policy function and $\mathcal{D}$ represents the training dataset. This pretraining process establishes robust semantic priors for subsequent optimization stages.

Building upon~\cite{ji2024regret}, the second phase implements a comprehensive MDP-based framework. The optimization objective is formulated as:

\begin{equation}
\label{eq:rl_objective}
J(\theta) = \mathbb{E}_{x \sim \mathcal{D}} \left[ \mathbb{E}_{l \sim p_l} \left[ \mathbb{E}_{s \sim \pi_{\theta}(\cdot|x,l)} [\mathcal{R}(x,s)] \right] \right],
\end{equation}
where we design a multi-component reward function balancing attack effectiveness with sequence diversity:

\begin{equation}
\label{eq:reward}
\mathcal{R}(x,s) = \alpha\mathcal{R}_{\text{atk}}(x,s) + \beta\mathcal{R}_{\text{div}}(x,l).
\end{equation}

To enhance training stability, we incorporate the reward smoothing mechanism proposed by~\cite{lee2023dreamsmooth}:

\begin{equation}
\label{eq:smooth_reward}
\tilde{\mathcal{R}}(x,s) = \frac{1}{k}\sum_{i=t-k+1}^t \mathcal{R}_i(x,s).
\end{equation}

This framework addresses the challenge of optimal sequence generation through three key innovations:
\pnum{1} Integration of supervised pretraining for establishing semantic priors
\pnum{2} Multi-component reward structure incorporating attack effectiveness and sequence diversity
\pnum{3} Advanced stability enhancement through temporal reward smoothing

The combination of these components enables efficient optimization of attack effectiveness while maintaining sequence naturalness, with the entropy-based reward mechanism ensuring robust performance throughout the training process.

\subsection{Specialized Sequence Generator}

The implementation of efficient zero-word-perturbation adversarial attacks necessitates a specialized generator architecture capable of precisely modeling the semantic relationships between textual content and emoji sequences. Building upon the foundational sequence-to-sequence framework established by~\cite{sutskever2014sequence} and~\cite{bahdanau2014neural}, we develop an enhanced architecture incorporating novel components specifically designed for emoji sequence generation.

A fundamental innovation in our approach lies in the construction of a unified vocabulary space that seamlessly integrates both textual and emoji tokens:

\begin{equation}
\label{eq:vocab}
\mathcal{V} = \mathcal{V}_t \cup \mathcal{V}_e,
\end{equation}
where $\mathcal{V}_t$ and $\mathcal{V}_e$ represent the text and emoji vocabularies, respectively. This unified representation enables coherent processing of both modalities during sequence generation. We enhance the standard attention mechanism through the introduction of a dynamic mask matrix $\mathbf{M}$, modulating information flow between modalities, with $\mathbf{M}_{ij} = 0$ for intra-modality attention and $\mathbf{M}_{ij} = \beta$ for cross-modality interactions.

The cornerstone of our architectural innovation is the \textcolor{teal}{\textbf{E}}moji \textcolor{teal}{\textbf{L}}ogits \textcolor{teal}{\textbf{P}}rocessor (ELP), implementing dynamic probability adjustment for token generation:

\begin{equation}
\label{eq:elp}
\mathbf{p}_{\text{out}} = \text{ELP}(\mathbf{p}_{\text{in}}) = \text{soft}(\mathbf{W}\mathbf{p}_{\text{in}} + \mathbf{b}).
\end{equation}

Our optimization framework incorporates multiple essential objectives. The semantic consistency component $\mathcal{L}_{\text{sem}} = -\log P(f_{\text{sen}}(s) = f_{\text{sen}}(x))$ enforces emotional alignment between generated sequences and input text, while the adversarial effectiveness measure $\mathcal{L}_{\text{adv}} = -\mathcal{R}(x,s)$ quantifies attack success. Following~\cite{zhang2024morl4pdes}, we employ an entropy-based diversity term $\mathcal{L}_{\text{div}} = -H(\pi_{\theta}(\cdot|x,l))$ to promote sequence variation. These components integrate into a unified objective function:

\begin{equation}
\label{eq:elp_loss}
\mathcal{L} = \mathcal{L}_{\text{sem}} + \lambda_1\mathcal{L}_{\text{adv}} + \lambda_2\mathcal{L}_{\text{div}},
\end{equation}
where hyperparameters $\lambda_1$ and $\lambda_2$ regulate the balance between competing optimization objectives. This specialized architecture, through its unified vocabulary representation, dynamic attention masking, and novel ELP component, enables effective generation of adversarial emoji sequences while maintaining semantic coherence with the original text content.

\section{Experiment}

To rigorously evaluate our proposed EmotiAttack framework, we designed a comprehensive experimental protocol utilizing two benchmark datasets: Go Emotion and Tweet Emoji. The Go Emotion dataset comprises emotion-labeled text instances across multiple fine-grained emotional categories, while the Tweet Emoji dataset consists of Twitter posts with associated emoji usage patterns. Both datasets provide diverse linguistic contexts, enabling thorough assessment of our framework's generalizability.

For target models, we employed two state-of-the-art transformer-based architectures: BERT and RoBERTa, both fine-tuned for their respective classification tasks. All experiments were conducted on an NVIDIA RTX 3090 GPU with 24GB VRAM. Our implementation utilizes PyTorch with the AdamW optimizer for gradient-based optimization.

To systematically investigate the relationship between search space dimensionality and attack performance, we evaluated four distinct search space configurations: top-1, top-3, top-15, and top-30. This experimental design enables precise characterization of the trade-off between attack effectiveness and computational efficiency. Throughout our evaluations, we maintain consistent preprocessing protocols and evaluation metrics to ensure fair comparison across different experimental conditions.
\begin{table*}[t]
\centering
\caption{Emoji-Attack's performance on RoBERTa and BERT models with perturbations of varying sizes.}
\label{tab:my-table-1}
\begin{tabular}{@{}c|c|c|c|c|c|c@{}}
\toprule
\textbf{Dataset}             & \textbf{Model}           & \textbf{Size} & \textbf{Pert. Rate} & \textbf{ASR (\%)} & \textbf{Avg. Time (s)} & \textbf{Avg. Query} \\ \midrule
\multirow{8}{*}{Go Emotion}  & \multirow{4}{*}{RoBERTa} & top1          & \textbf{0}          & 79.51             & 0.05                          & 1.00                      \\ \cmidrule(l){3-7} 
                             &                          & top3          & \textbf{0}          & 90.25             & 0.08                          & 1.40                      \\ \cmidrule(l){3-7} 
                             &                          & top15         & \textbf{0}          & 95.17             & 0.09                          & 3.01                      \\ \cmidrule(l){3-7} 
                             &                          & top30         & \textbf{0}          & 96.09             & 0.09                          & 3.77                      \\ \cmidrule(l){2-7} 
                             & \multirow{4}{*}{BERT}    & top1          & \textbf{0}          & 73.13             & 0.03                          & 1.00                      \\ \cmidrule(l){3-7} 
                             &                          & top3          & \textbf{0}          & 77.87             & 0.07                          & 1.46                      \\ \cmidrule(l){3-7} 
                             &                          & top15         & \textbf{0}          & 82.64             & 0.04                          & 4.10                      \\ \cmidrule(l){3-7} 
                             &                          & top30         & \textbf{0}          & 87.52             & 0.12                          & 8.10                      \\ \midrule
\multirow{8}{*}{Tweet Emoji} & \multirow{4}{*}{RoBERTa} & top1          & \textbf{0}          & 71.54             & 0.06                          & 1.00                      \\ \cmidrule(l){3-7} 
                             &                          & top3          & \textbf{0}          & 86.78             & 0.06                          & 1.46                      \\ \cmidrule(l){3-7} 
                             &                          & top15         & \textbf{0}          & 92.56             & 0.07                          & 2.92                      \\ \cmidrule(l){3-7} 
                             &                          & top30         & \textbf{0}          & 95.00             & 0.08                          & 3.85                      \\ \cmidrule(l){2-7} 
                             & \multirow{4}{*}{BERT}    & top1          & \textbf{0}          & 40.46             & 0.06                          & 1.00                      \\ \cmidrule(l){3-7} 
                             &                          & top3          & \textbf{0}          & 59.16             & 0.07                          & 2.08                      \\ \cmidrule(l){3-7} 
                             &                          & top15         & \textbf{0}          & 84.74             & 0.08                          & 4.89                      \\ \cmidrule(l){3-7} 
                             &                          & top30         & \textbf{0}          & 90.58             & 0.09                          & 6.62                      \\ \bottomrule
\end{tabular}%
\end{table*}

\subsection{Attack Performance Analysis}

Table~\ref{tab:my-table-1} presents the performance metrics of our EmotiAttack framework across varying search space configurations. The most notable characteristic is the consistently maintained 0\% perturbation rate across all experimental settings, underscoring our framework's fundamental advantage: preserving complete textual integrity of the original content while achieving effective model manipulation.

Our approach demonstrates remarkably high attack success rates (ASR) across both datasets. For the RoBERTa model on Go Emotion, we achieve ASR values ranging from 79.51\% with top-1 constraints to 96.09\% with top-30 search space. Similarly, on Tweet Emoji, RoBERTa exhibits vulnerability with ASR values from 71.54\% to 95.00\%. The BERT model shows substantial susceptibility as well, with ASR values ranging from 73.13\% to 87.52\% on Go Emotion and from 40.46\% to 90.58\% on Tweet Emoji.

The computational efficiency of our framework is particularly noteworthy, with average processing time per sample consistently below 0.12 seconds across all configurations. Query requirements remain minimal, with average query counts ranging from 1.00 to 8.10, significantly outperforming traditional adversarial attack methods that typically require substantially higher query volumes.

The results reveal a clear correlation between search space size and attack performance. As the search space expands from top-1 to top-30, we observe monotonic improvements in ASR across all experimental configurations. This scalable configurability enables flexible adjustment based on specific operational requirements. For instance, on the Tweet Emoji dataset with the BERT model, expanding from top-1 to top-30 increases ASR from 40.46\% to 90.58\%, while only modestly increasing the processing time from 0.06 to 0.09 seconds and queries from 1.00 to 6.62.

Interestingly, RoBERTa consistently exhibits higher vulnerability compared to BERT, particularly at smaller search space configurations. This differential susceptibility could be attributable to their distinct pre-training objectives and contextual encoding mechanisms.

\subsection{Performance Evaluation on State-of-the-Art Large Language Models}

\begin{table*}[t]
\centering
\caption{Emoji-Attack's performance on LLMs with perturbations of varying sizes.}
\label{tab:my-table-2}
\resizebox{\textwidth}{!}{%
\begin{tabular}{@{}c|c|c|c|c|c|c@{}}
\toprule
Dataset                      & Size  & Qwen2.5-7b-Instruct & Llama3-8b-Instruct & GPT4o   & Claude3.5Sonnet & Gemini-Exp-1206 \\ \midrule
\multirow{2}{*}{Go Emotion}  & Top15 & 89.00\%             & 80.37\%            & 75.00\% & 76.85\%         & 77.81\%         \\ \cmidrule(l){2-7} 
                             & Top30 & 92.66\%             & 85.68\%            & 79.00\% & 82.32\%         & 83.28\%         \\ \midrule
\multirow{2}{*}{Tweet Emoji} & Top15 & 92.32\%             & 87.10\%            & 81.00\% & 77.49\%         & 86.82\%         \\ \cmidrule(l){2-7} 
                             & Top30 & 95.34\%             & 90.12\%            & 86.00\% & 81.99\%         & 90.35\%         \\ \bottomrule
\end{tabular}%
}
\end{table*}

While traditional NLP models represent a significant application domain for adversarial attacks, recent advancements in large language models (LLMs) have fundamentally transformed the natural language processing landscape. To comprehensively evaluate our EmotiAttack framework's generalizability and practical significance, we conducted extensive experiments targeting state-of-the-art LLMs, including both open-source models (Qwen2.5-7b-Instruct, Llama3-8b-Instruct) and proprietary systems (GPT4o, Claude3.5Sonnet, Gemini-Exp-1206).

Table \ref{tab:my-table-2} presents the attack success rates across five distinct LLM architectures using our zero-word-perturbation approach. We evaluated two search space configurations—top-15 and top-30—to investigate the relationship between search space dimensionality and attack effectiveness in the LLM domain. The results demonstrate a remarkable finding: contemporary LLMs exhibit substantial vulnerability to our emoji-based adversarial attack methodology.

The experimental outcomes reveal several notable patterns. First, our approach achieves exceptional attack success rates across all evaluated LLMs, with ASR values ranging from 75.00\% to 92.66\% on the Go Emotion dataset and from 77.49\% to 95.34\% on the Tweet Emoji dataset. These high success rates underscore a fundamental vulnerability in current LLM architectures that persists despite their significantly larger parameter counts and more sophisticated pre-training regimes compared to traditional NLP models.

Among the evaluated models, Qwen2.5-7b-Instruct exhibits the highest susceptibility to our attack methodology, with ASR values reaching 92.66\% and 95.34\% under top-30 search space configuration for Go Emotion and Tweet Emoji datasets, respectively. Notably, even the most robust model in our evaluation, GPT4o, demonstrates substantial vulnerability with ASR values of 79.00\% and 86.00\% under the same experimental conditions.

The consistent vulnerability pattern across diverse LLM architectures—spanning different model families, parameter scales, and training methodologies—indicates a systemic weakness in how current language models process and interpret emoji sequences in conjunction with textual content. This finding has profound implications for LLM deployment in security-critical applications, particularly as these models increasingly serve as fundamental components in automated decision systems.

The observed vulnerability becomes particularly concerning when considering the rapid proliferation of LLM-based applications across critical domains including content moderation, information retrieval, and automated decision-making. Our results suggest that malicious actors could potentially manipulate model outputs through strategically positioned emoji sequences without modifying the underlying textual content, thereby circumventing traditional content filtering mechanisms that focus primarily on textual perturbations.

Furthermore, the scalable configurability of our attack methodology, demonstrated through the consistent performance improvements when expanding from top-15 to top-30 search space, highlights the framework's adaptability across different operational constraints. This configurability enables precise calibration of attack parameters to target specific model architectures effectively.

The demonstrated effectiveness of EmotiAttack against state-of-the-art LLMs represents a significant contribution to understanding model robustness in the evolving NLP landscape. By exposing a critical vulnerability that persists across both traditional models and cutting-edge LLMs, our findings underscore the urgent need for developing more robust defense mechanisms that specifically address emoji-based manipulation vectors.

\section{Conclusion}
In this paper, we introduce Emoti-Attack, a Zero-Perturbation Adversarial
Attacks on NLP Systems via Emoji Sequences. It can attack the small model and LLMs

\bibliography{references}
\bibliographystyle{IEEEtran}

\newpage

 




\vfill

\end{document}